\documentclass[journal]{IEEEtran}
\usepackage{caption}
\usepackage{tikz}
\usepackage{amssymb}
\usepackage{booktabs}
\usepackage{amsmath}
\usepackage[linesnumbered,ruled]{algorithm2e}
\usepackage{algpseudocode}
\usepackage{caption}
\usepackage{tabulary}
\newcommand\Tstrut{\rule{0pt}{2.6ex}}     

\usepackage{array,booktabs,tabularx}

\usepackage{cite}

\usepackage{graphicx}
\usepackage{amssymb}
\usepackage{floatrow}
\usepackage{amsmath}
\usepackage{lipsum}
\usepackage{multicol}
\usepackage{tikz}
\newcommand{\nonl}{\renewcommand{\nl}{\let\nl\oldnl}}

\usetikzlibrary{arrows,shapes,positioning,shadows,trees}

\tikzset{
  basic/.style  = {draw, text width=2cm, drop shadow, font=\scriptsize, rectangle},
  root/.style   = {basic, rounded corners=2pt, thin, align=center,
                   fill=white!30},
  level 2/.style = {basic, rounded corners=4pt, thin,align=center, fill=white!60,
                   text width=5em},
  level 3/.style = {basic, thin, align=left, fill=white!60, text width=5em}
}

\newcommand{\CASE}[1]{\STATE \textbf{case} #1\textbf{:} \begin{ALC@g}}
\newcommand{\ENDCASE}{\end{ALC@g}}

\newcommand{\DEFAULT}{\STATE \textbf{default:} \begin{ALC@g}}
\newcommand{\ENDDEFAULT}{\end{ALC@g}}
\newcommand{\DEFAULTLINE}[1]{\STATE \textbf{default:} }

\ifCLASSINFOpdf
\else
\fi
\begin{document}
%
\title{IoT-based Route Recommendation for an Intelligent Waste Management System}
%
%

\author{Mohammadhossein Ghahramani,~\IEEEmembership{Member,~IEEE,}
           Mengchu Zhou,~\IEEEmembership{Fellow,~IEEE,}\\
           Anna Molter,
          and Francesco Pilla
\thanks{This work was supported by Science Foundation Ireland under Grant 16/IA/4610. The authors would like to acknowledge both Bigbelly and Futurestreet for providing the data used in this work. For the purpose of Open Access, the author has applied a CC BY public copyright licence to any Author Accepted Manuscript version arising from this submission.}
\thanks{M. Ghahramani is with the Spatial Dynamics Lab, University College Dublin, Ireland, (e-mail: sepehr.ghahramani@ucd.ie).}
\thanks{M. C. Zhou is with the Helen and John C. Hartmann Department of Electrical and Computer Engineering, New Jersey Institute of Technology, Newark, NJ 07102, USA (e-mail: zhou@njit.edu).}
\thanks{A. Molter is with the Spatial Dynamics Lab, University College Dublin, Ireland, (e-mail: anna.molter@ucd.ie).}
\thanks{F. Pilla is with the Spatial Dynamics Lab, University College Dublin, Ireland, (e-mail: francesco.pilla@ucd.ie).}
}
%
%

\markboth{}%
{Shell \MakeLowercase{\textit{et al.}}: Bare Demo of IEEEtran.cls for IEEE Journals}
%



\maketitle

\begin{abstract}
The Internet of Things (IoT) is a paradigm characterized by a network of embedded sensors and services. These sensors are incorporated to collect various information, track physical conditions, e.g., waste bins' status, and exchange data with different centralized platforms. The need for such sensors is increasing; however, proliferation of technologies comes with various challenges. For example, how can IoT and its associated data be used to enhance waste management? In smart cities, an efficient waste management system is crucial. Artificial Intelligence (AI) and IoT-enabled approaches can empower cities to manage the waste collection. This work proposes an intelligent approach to route recommendation in an IoT-enabled waste management system given spatial constraints. It performs a thorough analysis based on AI-based methods and compare their corresponding results. Our solution is based on a multiple-level decision-making process in which bins' status and coordinates are taken into account to address the routing problem. Such AI-based models can help engineers design a sustainable infrastructure system.
\makeatletter{\renewcommand*{\@makefnmark}{}
\footnotetext{}\makeatother}

\end{abstract}


%
\IEEEpeerreviewmaketitle

\section{Introduction}\label{section.intro}
%
%
%
%

\IEEEPARstart{T}{he} world's population living in urban areas is expected to continue to grow at a fast pace. Such population growth can lead to an increasing waste generation. Today, big cities worldwide face various waste management challenges due to rapid growth in population and consumption. Generally speaking, waste management consists of different procedures like garbage collection, transport, processing, waste disposal, and monitoring \cite{Ahmad2020Optimal1}. The challenges in this field include insufficient infrastructure for waste collection, lack of financial resources, health issues, and proper waste management planning \cite{Imran2020Quantum}. In this work, we deal with waste collection and route optimization. A waste management process is explicitly identified as a service. It consumes a significant portion of a municipality's operating budget. A considerable amount of this budget goes on garbage collection and transportation. The scenario seems to be an unreasonable wastage of resources if bins are collected while partially filled up. To deal with this situation, bins should be built on a microcontroller-based platform to be monitored effectively. In this way, the information about the status of bins (i.e., the level of garbage in bins) can be obtained before dispatching garbage collector vehicles for garbage collection. The conventional way of collecting garbage is a cumbersome procedure requiring much individual effort, time, and cost \cite{Wang2021Urban}. Developing new enabling technologies can help alleviate the situation. Artificial Intelligence (AI) has made its way into different areas, e.g., engineering \cite{Xiong2020Wafer,Sepehr2021SpatioIOT}, urban planning \cite{Sepehr2021Tales}, and management \cite{Sepehr2021Carbon}. AI-based models are incorporated in different fields of study, such as manufacturing, to solve various problems like process planning, route optimization, and extracting insights from sensor data \cite{Catarinucci2020IoT,Ghahramani2020AI,Ghahramani2020Urban,Ghahramani2019Extracting}. AI and IoT technologies can pave the way to develop an efficient system and automate waste management practices. They offer new strategies for reducing the cost, and the complexity of waste management \cite{Sheng2020Internet}.

AI-based approaches like Genetic Algorithm (GA) can be adopted to increase efficiency and reduce cost \cite{ghahSMC}. Their applications to waste management can make garbage collection processes smart. Such methods can substitute traditional ones for planning routes for waste collection while using fewer resources in transportation. Route planning in the context of waste management is defined as the process of identifying and evaluating the feasibility of routes for collecting garbage efficiently. This process includes different phases, e.g., identifying pickup locations, clustering locations/bins, measuring the amounts of waste to be collected, and planning a route \cite{Ahmad2020Optimal}. Proposing an efficient and practical waste collection approach is consistent with the characteristics of a given problem. In this work, we develop our solution based on the inherent properties of the concerned problem using Evolutionary Algorithms. The problem is formulated and modeled based on two different scenarios, i.e., discrete optimization \cite{Hou2017Pareto} and continuous one \cite{Aldhafeeri2019Brain}. Spatial constraints are integrated into the implemented models. They are employed to address proximity issues to be explained later. First, we explain how a routing problem is modeled based on both approaches. Then, we compare them and show the former (i.e., the proposed discrete optimization model) can provide a much better solution than the latter. Finally, different discrete algorithms are compared with the proposed discrete model to validate the result.

All bins are built on a microcontroller-based platform, embedded with sensors informing about the level of garbage using an Application Programming Interface (API). Our solution includes several phases. The first phase deals with defining different fullness levels, e.g., 60\% capacity of a given bin. Whenever bins come to this predefined level, they are given a priority upon their collection. Moreover, those bins whose capacity does not reach a minimum level (e.g., 20\% capacity) are not considered in the second phase of the algorithm. It should be mentioned that considering this flexibility can reduce the complexity of the algorithm significantly. The status of bins is obtained by querying the interface using API. The second phase concerns planning the best path given the bins' status determined in the prior phase. Dealing with spatial optimization problems is more challenging than non-spatial ones. Topological constraints should be considered when solving a spatial optimization problem; in other words, the physical properties of spatial points and their neighbors should be considered. The main contributions of an efficient spatial optimization method include computational efficiency, flexibility, and optimization quality. In line with the mentioned goals, we propose an approach that improves computational efficiency and achieves much higher optimization quality than existing ones. The contributions of this work are as follows:

\begin{enumerate}
\item An optimized IoT-based model including intelligent vehicle routing strategies coupled with spatial constraints is proposed to offer promising search capability in a discrete and continuous domain for waste management.

\item The model can maintain diverse solutions given different decision-making constraints for sustainability concerns. 
\end{enumerate}

The remainder of this paper is organized as follows: some related work about the models that have been designed for waste collection is described in Section \ref{RelatedWorks}; both implemented models are discussed in Section \ref{Models}; the experimental settings and the comparison results are shown in Section \ref{result}; and the future work and conclusions are presented in Section \ref{conclusion}.

\section{Related Work}\label{RelatedWorks}
Recent studies have been focused on different ways to exploit the opportunities offered by AI and machine learning due to the limitations of conventional computational approaches \cite{Chang2015Sus}. In environmental engineering, leveraging IoT sensors has enabled access to data that can be traced throughout various planning phases, from higher-level strategic planning processes to lower-level operational planning. AI-based techniques such as Artificial Neural Network, multi-layer perception, Adaptive Neuro-Fuzzy Inference System, and Evolutionary Algorithms have been implemented to address air pollution and emissions reduction \cite{Gu2018Recurrent, Abdulla2018Integrated,Liu2020Parallel}. Moreover, different technologies, such as cloud and edge computing \cite{Wan2020Cloud, Ghahramani-4,SHI1,Xia2020QoE,Deng2020Edge}, are also embedded in today's urban infrastructures \cite{zanella1,ghah1,ghah2,ghah3}. These technologies are used to store and process data from almost anywhere. They are being adopted across various enterprises to improve operations and develop better analytics to enhance decision-making ability. Given these technologies, intelligent data-driven mechanisms can be developed to enable authorities to manage waste effectively.

Broadly speaking, route optimization is referred to as an NP-Hard problem. Therefore, most studies in this context are based on heuristic and meta-heuristic algorithms \cite{Hawbani2021Novel,Lee2020Resource,Ahmad2021Heuristic,Fanti2021Route,Jin2021Planning}. However, meta-heuristics can provide more robust, effective, and cost-efficient solutions to multi-objective problems. The quality of solutions provided by these algorithms is much higher than those obtained by conventional heuristic ones \cite{Lei2019Two}. Hence, meta-heuristics have received considerable attention for solving problems. Wy et al. \cite{Wy2013waste} develop a heuristics method to model a rollon–rolloff waste collection vehicle routing problem. To do so, they propose a neighborhood search-based iterative heuristic approach. In \cite{Mu2016Solving}, a parallel meta-heuristic approach based on simulated annealing (SA) is proposed to incorporate asynchronous and synchronous Markov chains. A residual capacity and radical surcharge algorithm have been considered to generate an initial solution for the model; then, local search methods are implemented to optimize the obtained solution. These models can be computationally expensive since heuristic and meta-heuristic algorithms are integrated. Wang et al. have introduced a memetic algorithm with competition to solve the routing problem \cite{Wang2019memetic}. Their solution is based on a k-nearest neighbor approach coupled with a SA strategy. First, they have implemented a permutation-based method to encode the solution. Then, an effective decoding method has been constructed. Finally, SA has been used to find optimal routes. Different variants of the classical SA algorithm have also been adopted for route optimization in \cite{Tirkolaee2019Developing, Tirkolaee2016Solving}. However, a sequence of collection tasks has not been considered in these studies. To alleviate this concern, an optimization problem is solved by using meta-heuristics such that the number of collection trips is minimized \cite{Zhang2020Solution}. The work \cite{Zhang2020Solution} provides a theoretical model for a routing problem with the characteristics of full loads and multiple trips. A traveling salesman problem \cite{Huang2020Niching}, as a combinatorial optimization problem, has been considered to model vehicle routing problems \cite{Wang2016Multiobjective, Kim2015City}. For these problems, diverse optimal solutions should be provided. GA has been used in waste management planning to model waste accumulation, facility siting, and generation \cite{Toutouh2019Computational}. Amal et al. have proposed a GIS-based Genetic Algorithm for optimizing the route of solid waste collection. They use a modified version of the original Dijkstra algorithm in GIS to generate optimal solutions. They have also conducted a case study at Sfax city in Tunisia to validate the performance of their method \cite{Amal2018SGA}.

Several approaches to route planning for waste management have been proposed in the literature. Some of them are based on periodic collection \cite{Wy2013waste}; while in some others, the collection procedure is done within a specific time window \cite{Tirkolaee2019Developing}. Collection approaches can vary according to urban characteristics, i.e., residential or commercial/industrial areas. In residential areas, the collection is primarily on a door-to-door basis and carried out by vehicles passing along the streets to collect accumulated garbage from households. In commercial districts, trucks visit different locations, e.g., designated business locations like hotels and shopping malls or waste containers in industrial regions, to collect garbage. Despite the advantages of the existing models, they are not efficient if used to solve our concerned problem. They cannot appropriately model the associations among spatial objects. Effective solutions are highly dependent on the characteristics of problems to be solved. For example, we deal with proximity issues in this work. To address such concerns, spatial constraints should be integrated into our approach. This work aims to present an efficient approach given the problem characteristics while addressing all related concerns. It also compares its results with other methods.

\section{Waste Collection Models}\label{Models}
The proposed IoT-based model is based on integrated algorithms, including GA and an artificial neural network. The associated cost of the model is calculated by incorporating ANN into the model. Generally speaking, GA consists of different phases, i.e., parent selection, crossover, mutation, and creating the final population \cite{Ghahramani2020AI}. Parent selection is a vital phase, consisting of a finite repetition of various procedures, such as selecting parent strings, recombining strings, and mutation operations. The reproductive phase's goal is to choose cost-efficient individuals from the population and produce new offsprings for the next generation. An effective mechanism should be considered to deal with the exploration and exploitation of an algorithm and avoid premature convergence. The mentioned concerns can lead to a loss of diversity. An efficient solution should also eliminate the cost scaling issue and adjusts the selection procedure. All related concerns have been addressed in this work. We consider a selection pressure approach and adjust the balance between exploration and exploitation by recombining crossover operators to adjust their probabilities. Consequently, individuals are produced throughout the mating pool by establishing a hybrid roulette tournament pick operator. A discussion for determining the exploration and exploitation rate is presented throughout the paper. Given the proposed model, garbage collectors can be sent to respective locations to collect garbage based on all bins' status. The implemented approach is practicable for different route planning.

\subsection{Spatial Constraints}
\begin{figure}
  \includegraphics[width=\linewidth]{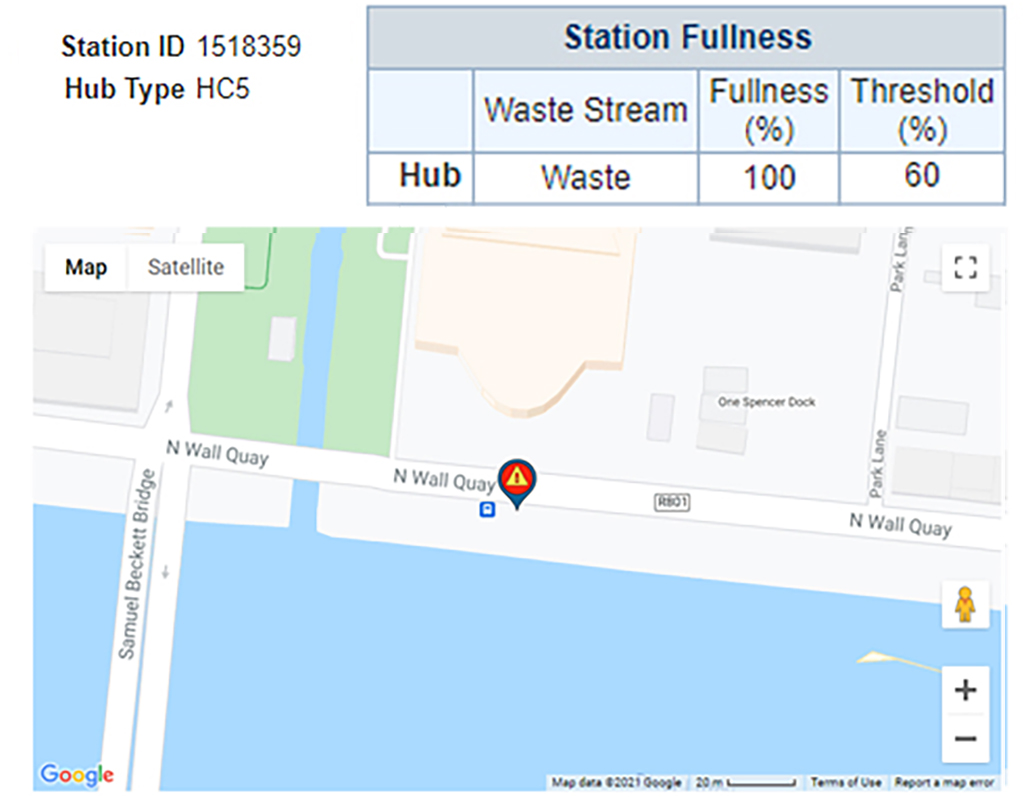}
  \caption{Querying API to check the status of waste bins.}
  \label{API}
\end{figure}

\begin{figure}
  \includegraphics[width=\linewidth]{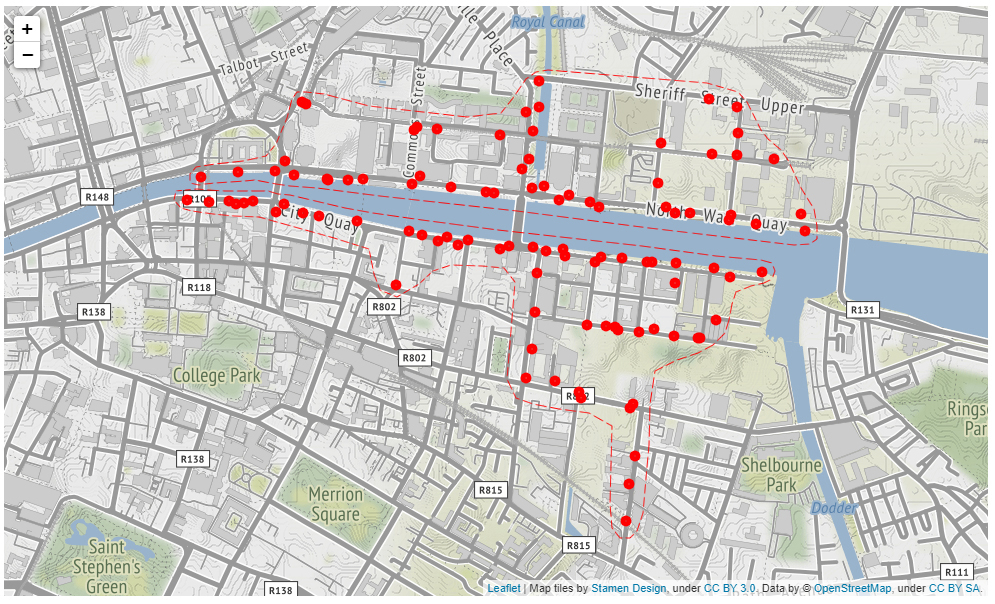}
  \caption{Distribution of bins in Docklands, Dublin.}
  \label{docklandsBins}
\end{figure}

The information about bins is obtained by querying API of a waste management company (i.e., Bigbelly bins and Futurestreet) based in Dublin, Ireland. The model interactively submits Hypertext Transfer Protocol (HTTP) requests, and the server returns responses. The response contains spatial and non-spatial information, coordinates of bins in Docklands (i.e., a vibrant area in Dublin), and their corresponding fullness status (Fig. \ref{API}). The parameters of HTTP requests such as API Token, Station IDs, StartTime, and EndTime are also contained. We deal with spatial characteristics and associated complexities in this work. Spatial analysis can be much more challenging than traditional data processing due to the complexity of possible patterns. In such an analysis, observations should be defined based on spatial relationships, e.g., spatial adjacency \cite{Guo2019Small}. Different methods (e.g., graph-based, grid neighbors, K-nearest neighbors, distance-based, and higher-order methods) define adjacency for a point or areal data. Some of these techniques are based on inter-point distances like k-nearest neighbors, while others are based on neighbors of areal units. As explained, an effective spatial analysis should be developed based on such proximity measures. However, defining adjacency criteria depends on the characteristics of a given data and relies on the spatial distribution of observations (e.g., bins together with all related features). The distribution of bins in Docklands (the area where our concerned bins are located) is demonstrated in Fig. \ref{docklandsBins}. As can be seen, a river runs through the area where the bins are located. In this scenario, defining a neighborhood list is challenging as finding shared boundaries among different areal units does not account for cross-area relationships segregated by bridges. Those units that are connected by bridges cannot be considered as neighborhood links. Therefore, we have implemented the neighborhood list used in the proposed algorithm based on a graph-neighbor approach to overcome such shortcomings. To that end, two different graph-based techniques, i.e., Delaunay triangulation and Gabriel graph methods \cite{Sun2010Edge}, are used. We have found that the former results in a more robust outcome. By defining this proximity measure, bins are automatically divided into two groups. As illustrated in Fig. \ref{docklandsBins}, two groups of bins can be separated automatically.

\subsection{Spatial Objective Function}
Suppose that, at $n$ spatial data points (bins) $b_i, i \in {1, 2, ..., n}$, we observe $y_i=y(b_i)$. $y(b_i)$ is the status of the $i^{\text{th}}$ bin. This vector can be defined as: 

\begin{equation}
Y=\begin{bmatrix}
         y_{1} ,   y_{2} , \ldots, y_{n}
        \end{bmatrix}^\top
  \end{equation}

Then, each $b_i$ is defined as: $b_i$ = (longitude; latitude; $Y_{i}$; $C_{i}$), including the coordinates of all bins in Docklands, $Y_{i}$ is the associated  percentage of fullness of the $i^{\text{th}}$ waste bin $b_i$, and $C_{i}$ represents the cluster that the $i^{\text{th}}$ bin belongs to. It is worth mentioning that each $b_{i}$ is implemented as a \textit{spatial data point}, consisting of different slots, such as projection-string, plot-order, and bounding-box. These features are used for projection purposes. The defined spatial data points are the objects of Coordinate Reference Systems. 

The spatial-weights-matrix $W$ is also defined by considering all nearby bins as:
\begin{equation}
W=\begin{bmatrix}
         w_{ij}
        \end{bmatrix}_{n \times n}\end{equation}
the weights between pairwise adjacent bins $b_i$ and $b_j$ satisfy:
\begin{equation}
w_{ij} > 0 \mbox { for any adjacent bins},\mbox{    and }w_{ii} {=} 0 \mbox { for any } i.
\end{equation}

Let $G = (V, E)$ be an  \textit{undirected weighted graph} with $n$ bins, where $V$ = \{$V_1, V_2, ..., V_n$\} is a set of spatial nodes, and $E$ is a matrix representing connecting edges (indicating the path between two nodes). i.e.,

\begin{equation}
E=\begin{bmatrix}
         E_{ij}
        \end{bmatrix}_{n \times n}\end{equation}
where $i$ and $j$ represent bins $i$ and $j$. The distance among bins is also considered as the edge's length (e.g., the weight value of edge $(i,j)$ is denoted as $d_{ij}$). The desired path (the algorithm's objective function) can be formulated as the permutation $\Theta$.

\begin{equation}
\min f(\Theta)=\sum\limits_{k=1}^{n-1} d_{\Theta(k)\Theta(k+1)}+d_{\Theta(n)\Theta(1)}
\end{equation}
where $\Theta(k)$ is the $k^{\text{th}}$ element of permutation $\Theta$.        

Based on the defined list of bins and their pairwise distances, the aim is to search for the shortest path in a weighted graph such that each bin is visited exactly once. This problem is an NP-hard optimization problem and computationally expensive since the number of permutations of $n$ nodes grows exponentially with $n$. Given the nature of this problem, it is challenging to propose a solution based on deterministic algorithms. Hence, we propose two different approaches based on meta-heuristic techniques to obtain near-optimal solutions. The implemented techniques are population-based models coupled with meta-heuristic search algorithms. Such approaches enable us to find a near-optimal solution within a reasonable time while avoiding the need for exhaustively exploring the input space. It should be mentioned that most studies have been focused on proposing solutions based on a continuous optimization domain. We aim to find an appropriate algorithm so that it can offer promising search capability and maintain diverse solutions. Since we deal with a combinatorial optimization problem, a GA is selected. It should be noted that some of the evolutionary algorithms, such as Particle Swarm Optimization (PSO) or covariance matrix adaptation evolutionary strategies, are originally proposed for a continuous search space \cite{Wang2020Modeling}.

GA includes a repetition of different operations sequences, i.e., parent selection, recombination, and mutation. We implement two variants of GA (with different genetic operators) in continuous and discrete cases. We show that the one conducting for the discrete space performs much better. The models treat all bins (chromosomes in GA terminology) as bit-strings. Both models rely on a population of individuals (candidate solutions) to explore a search space. These candidates are a set of chromosomes and encoded as strings. The models use an initial population and genetic operators such as crossover and mutation. These operations are implemented to produce a new generation by recombining a population's chromosomes. Then, fitter individuals are chosen given an objective function. The operators are also utilized for exploring and exploiting search space. It is of great importance to balance the exploration and exploitation of an algorithm. Exploring the input space is done to find an optimal solution by using a crossover operation. Besides, a mutation operator is considered to avoid premature convergence of the algorithm. We have controlled the exploration/exploitation level by a selection pressure parameter ($\beta$ in this work). This parameter $\beta$ is used in the parent selection phase. All the procedures, iteratively, are repeated until some termination criteria are met. The best solution (the one with the minimum cost when the algorithm ends) is then selected.

As explained, the goal is to find the best route for waste collection. The first step is to define an initial population. Then, dynamic crossover and mutation operators are developed. The defined spatial objective function is used to minimize the total traveled distance. The procedures are described for both scenarios next.

\subsection{Implemented Model: Discrete Scenario}

In this scenario, a discrete GA coupled with the defined spatial objective function is implemented. To that end, an initial population is randomly generated (given the algorithm's population size). The cost associated with each chromosome is measured based on the defined objective function. These costs should be minimized. The initial population is defined as individuals, including $n$-dimensional chromosomes. A path set ($\zeta$) can be defined as a permutation of $n$ bins.

  \begin{equation} 
  \zeta= (b_{1}, b_{2}, \ldots, b_{n})
  \end{equation}

 Algorithm \ref{GA1} represents the way in which the initial population is created. 

 \begin{algorithm}[!h]
\SetAlgoLined

    \SetKwProg{Fn}{Function}{}{}
    \SetKwInOut{Input}{Input}
    \SetKwInOut{Output}{Output}
    \Input{\text{Spatial-Data-Frame} $(b_1,b_2,..,b_n)$\\ $Dist \gets \text{distance function}$} 
    \Output{$cost\gets \text{fitness of each permutation} $}

	{$n \gets \text{number of bins}$}\\
	{$P \gets []$}\\
	{$\eta_{p} \gets \text{Size of population}$}\\
	
	{$L \gets 0$}\\
	{$ \text{Select bins given their level of fullness}$}\\
		\# \enspace \textit{CostFn} \\
    \For{$i\gets 1, ..., \eta_{p}$}{	    
    {$ P_i.position \gets \text{a random permutation}$}\\
    {$ P_i.cost \gets \text{calculate cost given:}$}\\
    \For{$i\gets 1, ..., n-1$}{
    {$ cost.append \gets \text{cost of each permutation:}$}\\
    {$ L=L+Dist(P_i , P_{i+1})$}
     }}\
	{$P \gets \text{Sort population ($P$);}$}\\
	
         {return $\text{initial individuals;}$}

    \caption{Pseudo-code for creating initial population given their cost (Discrete Scenario)}
    \label{GA1}

\end{algorithm}

Then, cost-efficient chromosomes from the population are sorted and selected. A parent selection phase is employed for creating a new population at each iteration. In the reproduction phase, permutative crossover and permutative mutation operators are implemented. These operations are to be explained later. All the procedures are repeated until a termination criterion is met. A feasible solution to the problem satisfies the discussed constraints. Let $F$ be the set of all solutions; the aim is to find an optimal one ($f^* \in F$) given the cost function. In another word, we are looking for a permutation $\Theta^*$ of $n$ bins ($b_1, b_2, ... , b_n$) such that:
 
 \begin{equation}
\sum\limits_{k=1}^{n} d(b_{\Theta^*(k)},b_{\Theta^*(k+1)}) \leq \sum\limits_{k=1}^{n} d(b_{\Theta(k)},b_{\Theta(k+1)})
\end{equation}

The cost of each solution is calculated by using an ANN model. These costs are used for different purposes like parent selection and then the solution. The objective in the parent selection phase is to select two solutions (with lower costs) from the population such that newly created offsprings would inherit their parents' characteristics. There are different methods to select parents, e.g., Random Selection, Rank Selection, Stochastic Universal Sampling (SUS), Tournament Selection, and Boltzmann Selection. We do not adopt the Random Selection since it has no selection pressure parameter. Rank Selection and SUS suffer from premature convergence. Applying these methods can easily lead to a local optimum. Instead, we have employed Boltzmann Selection. This method is inspired by Simulated Annealing and can help maintain a good diversity. The probability of a solution being chosen is measured given the below Boltzmann probability:

   \begin{equation} 
  p_{(i)}= \frac{e^{-\beta L_{i}}}{\sum_{k=1}^{\eta_{p}} e^{-\beta L_{k}}}
  \end{equation}
where $\eta_{p}$ is the size of the initial population, and $L$ is the defined cost function. $\beta$ is the selection pressure. Parents are determined according to probabilities, which are proportional to the costs calculated earlier. In other words, solutions with a lower cost are more likely to be selected than those with a greater one. It is worth mentioning that $\beta$ is selected, such that $\sum_{i\in H}^{} p_{(i)}=0.7$, where $H$ is the set of half of the best solutions (population is sorted according to their cost values, and $\eta_{p}/2$ of them are selected). Consequently, the Roulette Wheel mechanism is implemented for sampling. This process is repeated until a predefined number of parents are selected. Therefore, solutions with the greatest cost have a minimal chance of being chosen. These procedures are described in Algorithm \ref{GA2}.

After selecting parents, a crossover operation should be employed. On this basis, the chromosomes of selected parents are combined to create new offspring. We deal with a combinatorial problem; hence, traditional methods (like one-point, two-point, and uniform crossovers) for conducting the crossover operation are not suitable. Two cut points on each pairwise parent have been considered in order to create a permutative crossover. The portion after each cut point has been selected and exchanged (the selected bit strings of the first parent are mapped onto the other parent's string). All bits are then checked to ensure there are no conflict bits to guarantee each chromosome is permutation without repetition. In this way, new offsprings are created. To avoid being trapped in a local minimum, a mutation operator has been considered. It also maintains a good diversity in a newly generated population. Different types of mutation operations (e.g., insertion, inversion, and swap) have been considered in our model. Generally, one or more bits of chromosomes are replaced, such that generated offsprings also maintain permutation patterns. The pseudo-code realizing the algorithm operators is given in Algorithm \ref{GA3}.

As stated, the goal is to minimize cost function $L$. To that end, an Artificial Neural Network (ANN) has been integrated into the model. The model gets the objective function (i.e., $L$) as an input. Then, the ANN is used to measure corresponding cost values in different iterations. Iteratively, different solutions (permutations) are generated and evaluated by GA's operations. A Levenberg-Marquardt training algorithm consisting of two layers (15 neurons in the hidden layer) is utilized. It enjoys adaptive weights with full connectivity among neurons in the input and hidden layers. All costs are calculated, and the path is selected such that the corresponding cost is minimized.

 \begin{algorithm}[!h]
\SetAlgoLined

    \SetKwProg{Fn}{Function}{}{}
    \SetKwInOut{Input}{Input}
    \SetKwInOut{Output}{Output}
    \Input{\text{CostFn,} \\ $P \gets \text{Initial Population}$} 
    \Output{ \text{Optimal Path} }

	{$I^{Max} \gets \text{Maximum Iterations}$}\\
	{$ \theta \gets \text{Crossover rate}$}\\
	{$\eta_{c} \gets \text{Size of crossover (based on $\theta$)}$}\\
	{$\eta_{m} \gets \text{Size of mutation}$}\\
	
    \For{$i\gets 1, ..., I^{Max}$}{	    
	\# \enspace \textit{Crossover operation} \\
	    \textit{ $Pr(s\in P)= \frac{exp^{(-\beta)* \frac{L_s}{LargestCost}}}{\sum_{k=1}^{\eta_{p}} exp^{(-\beta)* \frac{L_{k}}{LargestCost}}}$} \\
  
    \For{$i\gets 1, ..., \eta_{c} $}{	    
    \textit{Select two parents} ($\gamma$) \textit{based on the RW function;} \\ 
    \textit{Generate two offsprins} ($\alpha$); \\   	    
    {$ [\alpha(i,1).position,  \alpha(i,2).position]\gets \text{CrossoverFn($\gamma$1.position, $\gamma$2.position)}$}\\
    {$ [\alpha(i,1).cost,  \alpha(i,2).cost]\gets \text{CostFn($\gamma$1.position, $\gamma$2.position)}$}\\
        }\

    	\# \enspace \textit{Mutation operation} \\
   	 \For{$i\gets 1, ..., \eta_{m} $}{	    
 	   \textit{Select one parent} ($\gamma$) \textit{based on the RW function;} \\    	    
  	  {$ [\alpha(i).position]\gets \text{MutationFn($\gamma$.position)}$}\\
  	  {$ [\alpha(i).cost]\gets \text{CostFn($\gamma$.position)}$}\\
        }\ 	
 	{$P \gets \text{ [P, Offsprings, and Mutants];}$}\\
  	{$P \gets \text{ Sort and select first $\eta_{p}$ individuals;}$}\\
   	{$S^* \gets \text{P(1).position;}$}\\
   	{$Cost^*(i) \gets \text{P(1).cost;}$}\\        
        }\	
         {return $\text{optimal path;}$}

    \caption{Pseudo-code for finding an optimal path (Discrete Scenario)}
    \label{GA2}

\end{algorithm}

\begin{algorithm}[!h]
\SetAlgoLined

    \SetKwProg{Fn}{Function}{}{}
    \SetKwInOut{Input}{Input}
    \SetKwInOut{Output}{Output}
    \Input{\text{Two parents, }$(\gamma)$} 
    \Output{\text{Offsprings}}

	\Fn{$ [O1,O2]=CrossoverFn(\gamma1,\gamma2)$}{
		{$CrossoverMethod \gets \text{ \{One-Point Crossover\}}$}\\
return two offsprings;\\}\textbf{end}\

	\Fn{$ M=MutationFn(\gamma)$}{
Apply three mutation operators (Swap, Inversion, and Insertion);\\
{$Pr_{swap} \gets  \text{ probability of using Swap operator;}$}\\
{$Pr_{insertion} \gets  \text{ probability of using Insertion operator;}$}\\
{$Pr_{inversion} \gets  \text{ probability of using Inversion operator;}$}\\
{$Method \gets RW (Pr_{swap}, Pr_{insertion}, Pr_{inversion});$}

\Switch{$Method$}{
            \Case{1}{
                Swap($\gamma$)
            }
            \Case{2}{
                Insertion($\gamma$)
            }
            \Case{3}{
                Inversion($\gamma$)
            }
        }

return mutant;\\
}\textbf{end}\

    \caption{Pseudo-code for defining the algorithm's operators (Discrete Scenario)}
    \label{GA3}

\end{algorithm}

\subsection{Implemented Model: Continuous Scenario}
In this scenario, we implement GA that leverages random keys to encode solutions. Despite the method used above (where a stream of integers is used to represent the order in which bins are to be visited), in this model, each bit string in a chromosome is assigned with a random number drawn uniformly from $[0, 1)$ and sorted in ascending order. Hence, each solution has two different parts, i.e., an integer part and a fractional one. The former refers to what we used for permutation, while the latter indicates random key numbers assigned to bit strings. The initial population is generated by creating $n$ chromosomes, where $n$ is the number of bins. Then, a real number drawn randomly from [0, 1) is given to each bin. New individuals are spawned in the algorithm's reproduction phase given a predefined crossover rate and mutation one. The procedure realizing the initial population generation is presented in Algorithm \ref{GA4}.

 \begin{algorithm}[!h]
\SetAlgoLined

    \SetKwProg{Fn}{Function}{}{}
    \SetKwInOut{Input}{Input}
    \SetKwInOut{Output}{Output}
    \Input{\text{Spatial-Data-Frame} $(b_1,b_2,..,b_n)$\\ $Dist \gets \text{distance function}$} 
    \Output{$cost\gets \text{fitness of each solution} $}
    
		{$ \text{Select bins given their level of fullness}$}\\
	{$n \gets \text{number of bins}$}\\
	{$\eta_{p} \gets \text{Size of population}$}\\
	{$S \gets n \text{ uniformly random numbers in the interval (0,1)}$}
	
	\# \enspace \textit{Turn a real vector into a permutation} \\
	{$S \gets sorted(S)$}\\	
	{$L \gets 0$}\\
		\# \enspace \textit{CostFn} \\
    \For{$i\gets 1, ..., n-1$}{
    {$ cost.append \gets \text{cost of each permutation:}$}\\
    {$ L=L+Dist(S_i , S_{i+1})$}
     }\	
         {return $\text{solutions and costs;}$}

    \caption{Pseudo-code for creating initial population given their cost (Continuous Scenario)}
    \label{GA4}
  \end{algorithm} 

After the initial population is generated, different operators are performed for passing the best solutions in the population to the next generation based on an elitist strategy. First, two parents are selected, and a parametrized single crossover is considered for producing two new offsprings. In the defined crossover, one random position in the array of bits is selected and exchanged. As discussed earlier, after the initial population is generated, the parent selection operation should be conducted in the reproduction phase. The objective is to choose solutions with minimal cost. The cost function and the ANN used for calculating costs are both the same as those described in the previous section. The newly added offsprings inherit bits based on a probability measure discussed earlier. Second, a mutation operator is implemented to ensure a diverse population. It should be mentioned that each time a new solution is generated, it is compared to previous solutions in the population to make sure no duplicate individuals are maintained. The maximum number of iterations (the algorithm termination criterion) is also set to 1000. Algorithm \ref{GA5} realizes the elitist strategy in the random-key approach.

 \begin{algorithm}[!h]
\SetAlgoLined

    \SetKwProg{Fn}{Function}{}{}
    \SetKwInOut{Input}{Input}
    \SetKwInOut{Output}{Output}
    \Input{\text{CostFn,} \\ $P \gets \text{Initial Population}$} 
    \Output{ \text{Optimal Path} }

	{$i \gets 0$}\\
	{$ \text{Randomly initial population $(P_i)$}$}\\
	{$\text{Evaluate each solution in $P_i$}$}\\
	{$\eta_{p} \gets \text{Size of population}$}\\
	{$ \text{Repeat}$}\\
	
    \For{$j\gets 1, ..., \eta_{p}$}{	    
	{$ \text{Select two parents from $(P_i)$}$}\\
	    \textit{ $P_j=CrossoverFn(P_1,P_2)$} \\
	    \textit{ $P_j=MutationFn(P_j)$} \\
	     $P_{i+1} \gets P_j$ \\        
        }\	
        $i \gets i+1$\\
         {return $\text{optimal path;}$}

    \caption{Pseudo-code for finding optimal path (Continuous Scenario)}
    \label{GA5}

\end{algorithm}

\begin{figure}
  \includegraphics[width=\linewidth]{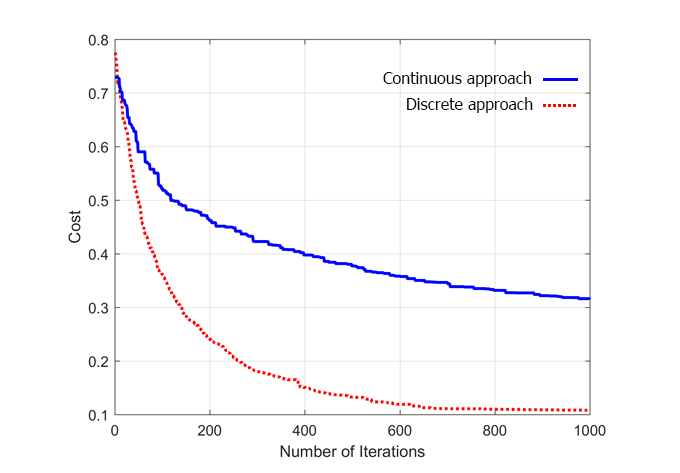}
  \caption{Comparing the costs of both models in different iterations.}
  \label{CostFunctionBins}
\end{figure}

\begin{figure}
  \includegraphics[width=\linewidth]{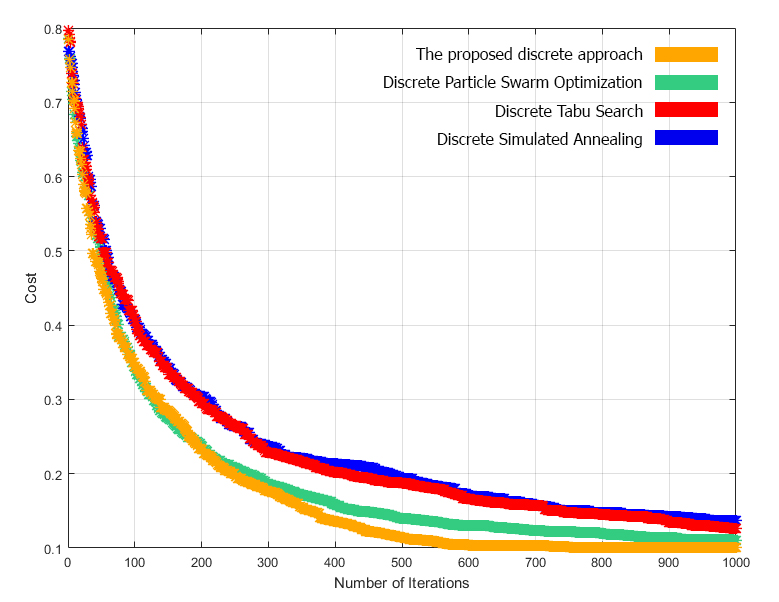}
  \caption{Comparing different discrete algorithms.}
  \label{CostFunctionAlgorithm}
\end{figure}

\begin{figure*}
  \includegraphics[width=\linewidth]{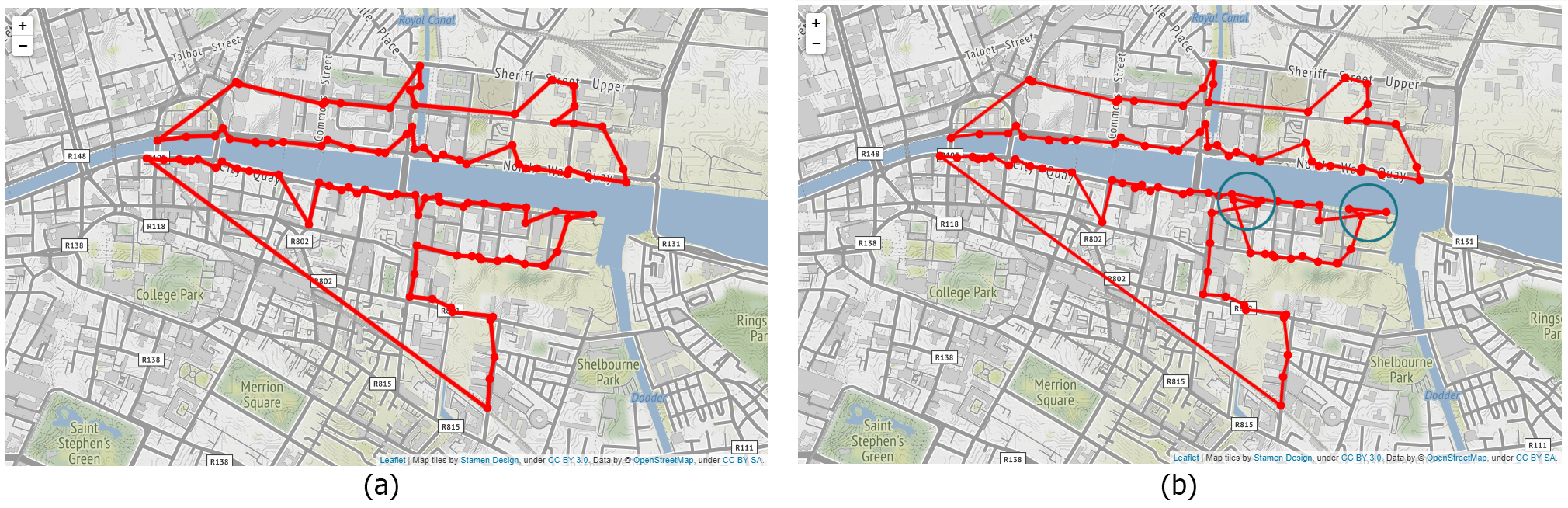}
  \caption{Results of two implemented models after 1000 iterations, i.e., a) optimal path given the disceret optimization model and b) optimal path given the continuous optimization model}
  \label{docklandsBins1}
\end{figure*}

\begin{figure*}
  \includegraphics[width=\linewidth]{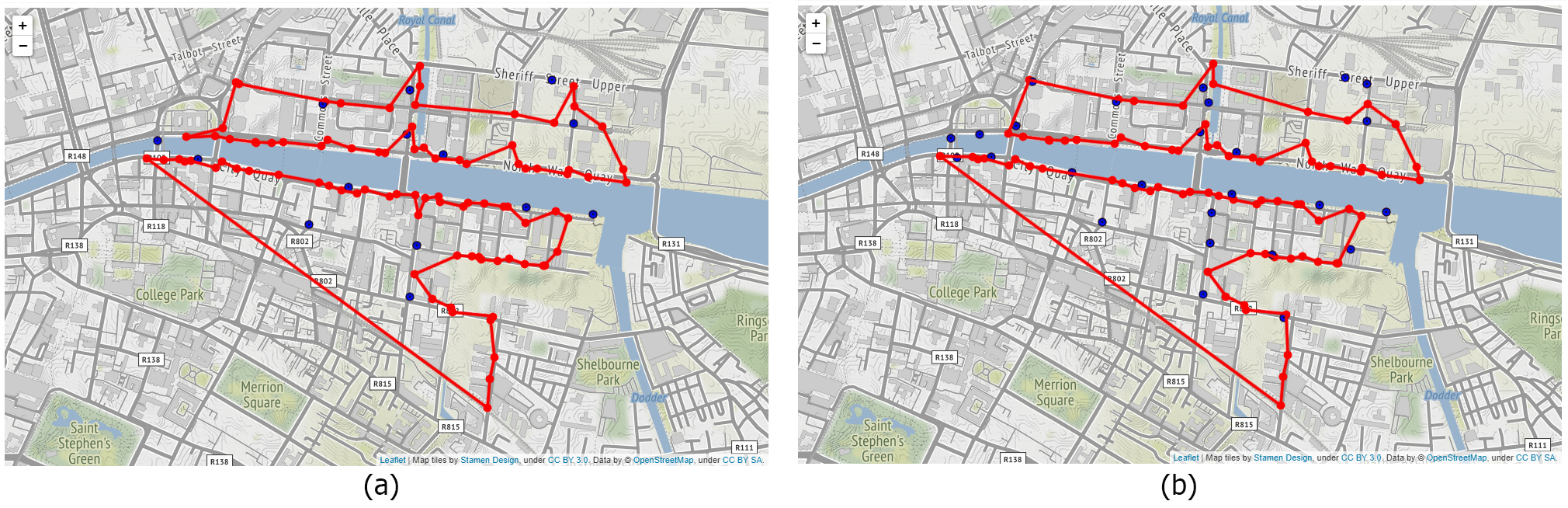}
  \caption{Results of the discrete model given two different settings after 1000 iterations: a) the level of bins' fullness is $20\%$,  b) the level of bins' fullness is $30\%$}
  \label{docklandsBins2}
\end{figure*}

\section{Results}\label{result}
In this work, we have proposed an optimized IoT-based waste collection model, including intelligent vehicle routing strategies coupled with spatial constraints. The goal is to present an efficient and practical waste collection approach given the characteristics of the concerned bins data. Dublin City Council has partnered with Big Belly Bins to install 110 smart bins across the Dublin Docklands area. All bins are built on a microcontroller-based platform, embedded with sensors informing about the level of garbage. These sensors provide real-time information showing the capacity of each of the bins. This allows us to see how much waste each bin contains in terms of percentages, i.e., 60\% or 80\%. The proposed IoT-based solution defines the level of fullness, and the optimal path is determined. We have proposed two different scenarios based on a hybrid approach. All experiments are conducted on an 18-core system with 192 GB memory. Two implemented hybrid meta-heuristic methods (integrated with ANN) evaluate various solutions to optimize the defined objective function. The volume of the dataset and the number of bins are considered for defining the initial population rate. The number of neurons is designated based on a trial and error method. Moreover, the volume of the concerned dataset has been considered in defining the initial population rate. As stated, the cost of each solution is calculated by using an ANN model. These costs have been used for different purposes like parent selection. To that end, a multilayer perceptron with Levenberg-Marquardt training algorithm has been used (since it converges faster and more accurately towards our problem) consisting of two layers of adaptive weights with full connectivity among neurons in the input and hidden layers. Note that ANN is only used for cost calculation; hence, we have not included much information about this phase. The model queries API, and the information of bins (including their locations and level of fullness) is obtained. The optimal route is then detected after a series of iterative computations (given the termination criterion, e.g., the number of iterations or computation time). Fig. \ref{CostFunctionBins} displays the cost values in each iteration for both models. As can be seen, the algorithm based on discrete optimization converges much faster than the continuous one. The experimental results and the impacts of different parameter settings for the discrete and continuous methods are presented in Tables \ref{Params1} and \ref{Params2}, respectively. We have also implemented several discrete evolutionary algorithms and compared their results with the proposed discrete approach. Fig. \ref{CostFunctionAlgorithm} reveals the results obtained from different discrete evolutionary algorithms and shows the effectiveness of our approach. The suggested paths resulting from both algorithms are demonstrated in Fig. \ref{docklandsBins1}. As can be seen in this figure, the second approach (Fig. \ref{docklandsBins1}(b)) is trapped in an optimal local solution. As discussed throughout the paper, the level of fullness of bins has also been taken into account. When collectors are dispatched to collect waste, the level of fullness of bins is monitored. Fig. \ref{docklandsBins2} illustrates the result given such consideration at a certain time. The level of fullness of bins depicted in blue in Fig. \ref{docklandsBins2}(a) is less than $20 \%$, and $30 \%$ for those in Fig. \ref{docklandsBins2}(b). Given all results, we can conclude that the proposed discrete model is superior to the continuous one.

\begin{table*}[!ht]
\caption{Results of the discrete optimization model given different parameter setting} 
\centering 
\scriptsize
\begin{tabular}{l c c c rrrrrrrrrrr} 
\hline\hline 
\Tstrut
 Crossover Rate  & Population Size & Neurons &\multicolumn{11}{c}{Corresponding cost for selected solution in different iterations}
\\ [0.5ex]
\hline 
\Tstrut
 & 
 &$6$ & $1.2031 $ & $1.2009$ & $1.1702$ & $1.1699$ & $1.1664$ & $...$ & $0.3212$ & $0.3212$ & $0.3178$ & $0.3178$ & $0.3178$  \\[-0.5ex]
{$0.6$} & {$70$}
& $10$ & $1.2102 $ & $1.1002$ & $1.1002$ & $1.1002$ & $1.0847$ & $...$ & $0.2915$ & $0.2905$ & $0.2833$ & $0.2830$ & $0.2765$  \\[-0.5ex]
 & 
 &$15$ & $1.3202 $ & $1.2723$ & $1.2721$ & $1.2621$ & $1.2586$ & $...$ & $0.3082$ & $0.3080$ & $0.3003$ & $0.3003$ & $0.3003$   \\[0.5ex]
\hline 
 & 
 &$6$ & $1.3124 $ & $1.3012$ & $1.2731$ & $1.1974$ & $1.1844$ & $...$ & $0.2523$ & $0.2475$ & $0.2378$ & $0.2378$ & $0.2370$  \\[-0.5ex]
{$0.6$} & {$80$}
& $10 $& $1.3012 $ & $1.1302$ & $1.1922$ & $1.1804$ & $1.1543$ & $...$ & $0.2814$ & $0.2805$ & $0.2805$ & $0.2805$ & $0.2711$  \\[-0.5ex]
 & 
 &$15$ & $1.2952 $ & $1.2653$ & $1.2653$ & $1.2653$ & $1.2641$ & $...$ & $0.3051$ & $0.3050$ & $0.2976$ & $0.2976$ & $0.2936$   \\[0.5ex]
\hline 
 & 
 &$6 $ & $1.4142 $ & $1.3839$ & $1.2943$ & $1.2602$ & $1.2461$ & $...$ & $0.2903$ & $0.2785$ & $0.2785$ & $0.2785$ & $0.2785$  \\[-0.5ex]
{$0.6$} & {$90$}
& $10$ & $1.3772 $ & $1.1372$ & $1.1362$ & $1.1361$ & $1.1347$ & $...$ & $0.2664$ & $0.2508$ & $0.2508$ & $0.2449$ & $0.2443$  \\[-0.5ex]
 & 
 &$15$ & $1.3912 $ & $1.3644$ & $1.2983$ & $1.2703$ & $1.2699$ & $...$ & $0.3001$ & $0.2751$ & $0.2707$ & $0.2707$ & $0.2700$   \\[0.5ex]
\hline 
 & 
 &$6 $& $1.0921 $ & $1.0622$ & $1.0613$ & $1.0613$ & $1.0509$ & $...$ & $0.2071$ & $0.2054$ & $0.2054$ & $0.1963$ & $0.1963$  \\[-0.5ex]
{$0.7$} & {$70$}
& $10$ & $1.0903 $ & $1.0741$ & $1.0608$ & $1.0573$ & $1.0491$ & $...$ & $0.2121$ & $0.2004$ & $0.1864$ & $0.1664$ & $0.1663$  \\[-0.5ex]
 & 
 &$15$ & $1.0873 $ & $1.0873$ & $1.0871$ & $1.0776$ & $1.0601$ & $...$ & $0.2215$ & $0.2215$ & $0.2035$ & $0.1914$ & $0.1843$   \\[0.5ex]
\hline 
 & 
 &$6 $& $1.0801 $ & $1.0801$ & $1.0773$ & $1.0654$ & $1.0491$ & $...$ & $0.2005$ & $0.2001$ & $0.1803$ & $0.1704$ & $0.1703$  \\[-0.5ex]
{$0.7$} & {$80$}
& $10 $ & $1.0691 $ & $1.0685$ & $1.0611$ & $1.0464$ & $1.0459$ & $...$ & $0.1822$ & $0.1820$ & $0.1820$ & $0.1614$ & $0.1612$  \\[-0.5ex]
 & 
 &$15 $& $1.0673$ & $1.0572$ & $1.0417$ & $1.0417$ & $1.0417$ & $...$ & $0.1726$ & $0.1710$ & $0.1703$ & $0.1702$ & $0.1702$   \\[0.5ex]
\hline 
 & 
 &$6 $& $1.0734$ & $1.0732$ & $1.0549$ & $1.0567$ & $1.0501$ & $...$ & $0.1819$ & $0.1810$ & $0.1753$ & $0.1715$ & $0.1711$  \\[-0.5ex]
{$0.7$} & {$90$}
& $10$ & $1.0601$ & $1.0579$ & $1.0449$ & $1.0447$ & $1.0441$ & $...$ & $0.1603$ & $0.1602$ & $0.1602$ & $0.1602$ & $0.1602$  \\[-0.5ex]
 & 
 &$15$ & $1.0626$ & $1.0567$ & $1.0549$ & $1.0402$ & $1.0402$ & $...$ & $0.1653$ & $0.1649$ & $0.1644$ & $0.1621$ & $0.1621$   \\[0.5ex]
\hline 
 & 
 &$6 $& $0.8832 $ & $0.8803$ & $0.8773$ & $0.8672$ & $0.8563$ & $...$ & $0.1321$ &$ 0.1304$ &  $0.1304$ &  $0.1301$ & $0.1301$  \\[-0.5ex]
{$0.8$} & {$70$}
& $10$ & $0.8737 $ & $0.8705$ & $0.8705$ & $0.8644$ & $0.8578$ & $...$ & $0.1316$ &$ 0.1294$ &  $0.1293$ &  $0.1281$ & $0.1280$  \\[-0.5ex]
 & 
 &$15 $& $0.8839 $ & $0.8812$ & $0.8752$ & $0.8638$ & $0.8557$ & $...$ & $0.1325$ &$ 0.1311$ &  $0.1311$ &  $0.1311$ & $0.1311$   \\[0.5ex]
\hline 
 & 
 &$6$ & $0.8124 $ & $0.8049$ & $0.7991$ & $0.7985$ & $0.7961$ & $...$ & $0.1104$ & $0.1101$ & $0.1095$ & $0.1095$ & $0.1094$ \\[-0.5ex]
\textbf{0.8} & \textbf{80}
& \textbf{10} & \textbf{0.7856} & \textbf{0.7819} & \textbf{0.7751} & \textbf{0.7685} & \textbf{0.7661} & $...$ & \textbf{0.1024} & \textbf{0.1024} & \textbf{0.1022} & \textbf{0.1021} & \textbf{0.1021}   \\[-0.5ex]
 &  
 &$15 $& $0.8094 $ & $0.8091$ & $0.7977$ & $0.7881$ & $0.7763$ & $...$ & $0.1088$ & $0.1081$ & $0.1075$ & $0.1075$ & $0.1068$   \\[0.5ex]
\hline 
 & 
 &$6$ & $0.8211 $ & $0.8026$ & $0.8026$ & $0.7988$ & $0.7981$ & $...$ & $0.1164$ & $0.1143$ & $0.1121$ & $0.1121$ & $0.1120$   \\[-0.5ex]
{$0.8$} & {$90$}
& 10 & $0.8193 $ & $0.8109$ & $0.7966$ & $0.7965$ & $0.7953$ & $...$ & $0.1105$ & $0.1105$ & $0.1086$ & $0.1086$ & $0.1086$   \\[-0.5ex]
 & 
 &$15 $& $0.8214 $ & $0.8052$ & $0.7888$ & $0.7885$ & $0.7883$ & $...$ & $0.1109$ & $0.1107$ & $0.1095$ & $0.1095$ & $0.1091$   \\[0.5ex]
\hline 
 & 
 &$6 $& $1.0874$ & $1.0838$ & $1.0775$ & $1.0663$ & $1.0661$ & $...$ & $0.1578$ & $0.1575$ & $0.1572$ & $0.1572$ & $0.1571$  \\[-0.5ex]
{$0.9$} & {$70$}
& $10 $& $1.0893$ & $1.0882$ & $1.0725$ & $1.0653$ & $1.0565$ & $...$ & $0.1562$ & $0.1561$ & $0.1561$ & $0.1559$ & $0.1559$   \\[-0.5ex]
 &  
 &$15$ & $1.0843$ & $1.0832$ & $1.0819$ & $1.0817$ & $1.0811$ & $...$ & $0.1616$ & $0.1610$ & $0.1608$ & $0.1608$ & $0.1608$   \\[0.5ex]
\hline 
 & 
 &$6$ & $1.0876$ & $1.0775$ & $1.0775$ & $1.0696$ & $1.0672$ & $...$ & $0.1408$ & $0.1407$ & $0.1407$ & $0.1388$ & $0.1388$   \\[-0.5ex]
{$0.9$} & {$80$}
& $10$ & $1.0802$ & $1.0801$ & $1.0795$ & $1.0763$ & $1.0697$ & $...$ & $0.1475$ & $0.1475$ & $0.1472$ & $0.1472$ & $0.1471$  \\[-0.5ex]
 &  
 &$15$ & $1.0818$ & $1.0808$ & $1.0784$ & $1.0678$ & $1.0645$ & $...$ & $0.1463$ & $0.1460$ & $0.1460$ & $0.1457$ & $0.1457$   \\[0.5ex]
\hline 
 & 
 &$6$ & $1.0823$ & $1.0821$ & $1.0805$ & $1.0766$ & $1.0751$ & $...$ & $0.1474$ & $0.1462$ & $0.1462$ & $0.1458$ & $0.1458$  \\[-0.5ex]
{$0.9$} & {$90$}
& $10$ & $1.0831$ & $1.0812$ & $1.0705$ & $1.0693$ & $1.0651$ & $...$ & $0.1398$ & $0.1398$ & $0.1397$ & $0.1395$ & $0.1394$  \\[-0.5ex]
 & 
 &$15$ & $1.0794$ & $1.0764$ & $1.0744$ & $1.0701$ & $1.0688$ & $...$ & $0.1418$ & $0.1416$ & $0.1415$ & $0.1412$ & $0.1411$   \\[0.5ex]
\hline 

\hline 
\end{tabular}
\label{Params1}
\end{table*}

\begin{table*}[!ht]
\caption{Results of the continuous optimization model given different parameter setting} 
\centering 
\scriptsize
\begin{tabular}{l c c c rrrrrrrrrrr} 
\hline\hline 
\Tstrut
 Crossover Rate & Mutation Rate & Population Size & Neurons &\multicolumn{9}{c}{Corresponding cost for selected solution in different iterations}
\\ [0.5ex]

\hline 
 & &$80$ 
 &$10$ & $1.1171$ & $1.1164$ & $1.1153$ & $1.1144$ & $...$ & $0.5383$ & $0.5383$ &  $0.5381$ & $0.5381$  \\[-0.5ex]
{0.6} & {$0.2$}&$100$
& $15$  & $1.1124$ & $1.1102$ & $1.1014$ & $1.1003$ & $...$ & $0.5221$ & $0.5217$ &  $0.5211$ & $0.5211$  \\[-0.5ex]
 & &$120 $
 &$20$  & $1.2134$ & $1.2117$ & $1.1137$ & $1.1034$ & $...$ & $0.5761$ & $0.5703$ &  $0.5703$ & $0.5703$   \\[0.5ex]
\hline 
 & &$80$ 
 &$10$  & $1.1106$ & $1.1104$ & $1.1093$ & $1.1054$ & $...$ & $0.4987$ & $0.4871$ &  $0.4854$ & $0.4853$  \\[-0.5ex]
{$0.6$} & {$0.3$}&$100$
& $15$ & $1.1102$ & $1.1088$ & $1.1075$ & $1.1028$ & $...$ & $0.4761$ & $0.4754$ &  $0.4701$ & $0.4680$  \\[-0.5ex]
 & &$120$ 
 &$20$  & $1.1113$ & $1.1096$ & $1.1095$ & $1.1053$ & $...$ & $0.4786$ & $0.4784$ &  $0.4784$ & $0.4783$   \\[0.5ex]
\hline 
 & &$80$ 
 &$10$  & $1.1152$ & $1.1098$ & $1.1082$ & $1.1066$ & $...$ & $0.4996$ & $0.4982$ &  $0.4982$ & $0.4981$  \\[-0.5ex]
{$0.6$} & {$0.4$}&$100$
& $15$ & $1.1151$ & $1.1062$ & $1.1043$ & $1.1003$ & $...$ & $0.4683$ & $0.4681$ &  $0.4677$ & $0.4674$  \\[-0.5ex]
 & &$120$ 
 &$20$  & $1.1161$ & $1.1102$ & $1.1084$ & $1.1068$ & $...$ & $0.4799$ & $0.4799$ &  $0.4799$ & $0.4799$   \\[0.5ex]
\hline 
 & &$80 $
 &$10$  & $0.8501$ & $0.8471$ & $0.8432$ & $0.8301$ & $...$ &$ 0.3919$ & $0.3886$ &  $0.3844$ & $0.3842$   \\[-0.5ex]
{$0.7$} & {$0.2$}&$100$
& $15$  & $0.8314$ & $0.8311$ & $0.8011$ & $0.7804$ & $...$ &$ 0.3654$ & $0.3634$ &  $0.3634$ & $0.3633$   \\[-0.5ex]
 & &$120$ 
 &$20$  & $0.8332$ &$ 0.8252$ & $0.7962$ & $0.7492$ & $...$ & $0.3493$ & $0.3486$ &  $0.3484$ & $0.3483$   \\[0.5ex]
\hline 
 & &$80$ 
 &$10$ & $0.7511$ & $0.7510$ & $0.7432$ & $0.7402$ & $...$ &$ 0.3413$ & $0.3411$ &  $0.3411$ & $0.3411$  \\[-0.5ex]
\textbf{0.7} & \textbf{0.3}&\textbf{100}
& \textbf{15}  & \textbf{0.7301} & \textbf{0.7255} & \textbf{0.7032} & \textbf{0.7024} & $...$ &\textbf{0.3215} & \textbf{0.3214} &  \textbf{0.3210} & \textbf{0.3210}  \\[-0.5ex]
 & &$120$ 
 &$20$  & $0.7577$ & $0.7431$ & $0.7355$ & $0.7207$ & $...$ &$ 0.3449$ & $0.3446$ &  $0.3443$ & $0.3443$  \\[0.5ex]
\hline 
 & &$80$ 
 &$10$ & $0.7985$ & $0.7742$ & $0.7651$ & $0.7554$ & $...$ &$ 0.3816$ & $0.3796$ &  $0.3776$ & $0.3776$  \\[-0.5ex]
{$0.7$} & {$0.4$}&$100$
& $15$  & $0.7921$ & $0.7523$ & $0.7501$ & $0.748$ & $...$ &$ 0.3843$ & $0.3842$ &  $0.3842$ & $0.3841$  \\[-0.5ex]
 & &$120 $
 &$20$  & $0.7935$ & $0.7746$ & $0.7653$ & $0.7512$ & $...$ &$ 0.3821$ & $0.3812$ &  $0.3812$ & $0.3810$   \\[0.5ex]
\hline 
 & &$80$ 
 &$10$  & $0.8321$ & $0.8244$ & $0.8201$ & $0.8107$ & $...$ &$ 0.3919$ & $0.3919$ &  $0.3919$ & $0.3918$    \\[-0.5ex]
{$0.8$} & {$0.2$}&$100$
& $15$ & $0.8113$ & $0.8018$ & $0.8003$ & $0.7982$ & $...$ &$ 0.3893$ & $0.3852$ &  $0.3842$ & $0.3838$   \\[-0.5ex]
 & &$120 $
 &$20$  & $0.8214$ & $0.8112$ & $0.8014$ & $0.8004$ & $...$ &$ 0.3896$ & $0.3896$ &  $0.3895$ & $0.3894$    \\[0.5ex]
\hline 
 & &$80$ 
 &$10$  & $0.8127$ & $0.8105$ & $0.8077$ & $0.8017$ & $...$ &$ 0.3931$ & $0.3911$ &  $0.3911$ & $0.3910$  \\[-0.5ex]
{$0.8$} & {$0.3$}&{$100$}
& $15$  & $0.8121$ & $0.8028$ & $0.7984$ & $0.7975$ & $...$ &$ 0.4013$ & $0.3991$ &  $0.3987$ & $0.3982$  \\[-0.5ex]
 & &$120 $
 &$20 $ & $0.8201$ & $0.8201$ & $0.8103$ & $0.7953$ & $...$ &$ 0.3924$ & $0.3924$ &  $0.3924$ & $0.3924$   \\[0.5ex]
\hline 
 & &$80 $
 &$10$  & $0.8654$ & $0.8444$ & $0.8259$ & $0.8233$ & $...$ &$ 0.3991$ & $0.3991$ &  $0.3990$ & $0.3989$   \\[-0.5ex]
{$0.8$} & {$0.4$}&$100$
& $15$ & $0.8626$ & $0.8611$ & $0.8527$ & $0.8422$ & $...$ &$ 0.3952$ & $0.3948$ &  $0.3947$ & $0.3943$  \\[-0.5ex]
 & &$120 $
 &$20$  & $0.8427$ & $0.8412$ & $0.8410$ & $0.8410$ & $...$ &$ 0.3892$ & $0.3887$ &  $0.3885$ & $0.3883$   \\[0.5ex]
\hline 

\hline 
\end{tabular}
\label{Params2}
\end{table*}

\section{Conclusions and Future Work}\label{conclusion}
We have proposed an IoT-based model coupled with spatial constraints to solve a route recommendation problem in this work. We have modeled and solved the concerned problem through two different scenarios, i.e., discrete and continuous optimization approaches. Both methods are validated on a case study, and the results are compared. Although most studies have focused on continuous optimization methods, we have shown that a discrete algorithm is more efficient in dealing with the problem. The latter can be easily trapped into local optima. Different types of operators (e.g., permutation-based crossover and mutation operations) are designed to maintain diverse candidates during a search operation. We have also implemented some specific measures to control the balance between the exploration and exploitation of both algorithms. An artificial neural network is utilized to calculate the associated cost in each iteration of the algorithms. Different discrete methods are also compared with our proposed discrete approach to validate the efficiency and effectiveness of the proposed model, and the results are illustrated. The main drawback of the state-of-the-art is that it cannot appropriately model the associations among spatial objects, consequently find an optimal route. The proposed model can enable us to find an optimal solution within a reasonable time. It should be mentioned that the model can recommend multiple routes. A driver can then select the most favorable route from the suggested ones. In this work, we only obtain information about bins that are located in Docklands. As a part of our future work, we plan to consider all bins in Dublin.



\ifCLASSOPTIONcaptionsoff
  \newpage
\fi

\begin{IEEEbiography}[{\includegraphics[width=1in,height=1.25in,clip,keepaspectratio]{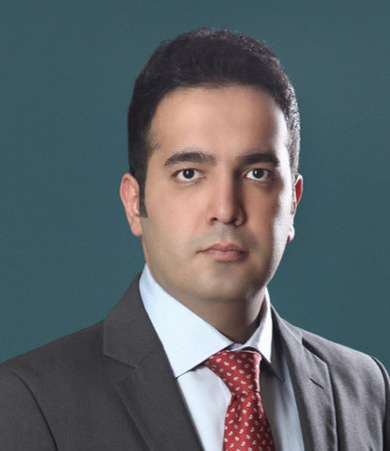}}]{Mohammadhossein Ghahramani}
(S'15-M'18) obtained his Ph.D. degree in Computer Technology and Application from Macau University of Science and Technology, Macau, in 2018. He is currently a Research Fellow at University College Dublin (UCD), Ireland. He also was a member of the Insight Centre for Data Analytics at UCD. His research interests include smart cities, smart manufacturing, machine learning, artificial intelligence, big data, and IoT. He has over ten peer-reviewed journal papers as a first author. Dr. Ghahramani has served the community by reviewing more than 100 papers for top journals in the last four years and has been active in organizing international conferences. He is a co-chair of IEEE SMCS Technical Committee on AI-based Smart Manufacturing Systems. \end{IEEEbiography}

\begin{IEEEbiography}[{\includegraphics[width=1in,height=1.25in,clip,keepaspectratio]{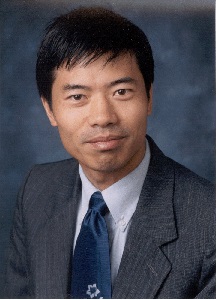}}]{MengChu Zhou}
(S'88-M'90-SM'93-F'03) received his Ph. D. degree from Rensselaer Polytechnic Institute, Troy, NY  in 1990 and then joined New Jersey Institute of Technology where he is now a Distinguished Professor. His interests are in Petri nets, automation, Internet of Things, and big data.  He has over 900 publications including 12 books, 600+ journal papers (500+ in IEEE transactions), 29 patents and 29 book-chapters. He is Fellow of IFAC, AAAS, CAA and NAI.
\end{IEEEbiography}

\begin{IEEEbiography}[{\includegraphics[width=1in,height=1.25in,clip,keepaspectratio]{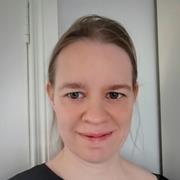}}]{Anna Molter} is a Post-Doctoral Research Fellow at University College Dublin (UCD), Ireland. Her research is about how the environment affects people's health. A lot of her work has focused on the effects of air pollution on children's health, but she is interested in a wide range of environmental issues, including both urban and rural problems. She likes to use tools such as apps and storymaps to communicate her findings with the public and to engage them in scientific research.
\end{IEEEbiography}

\begin{IEEEbiography}[{\includegraphics[width=1in,height=1.25in,clip,keepaspectratio]{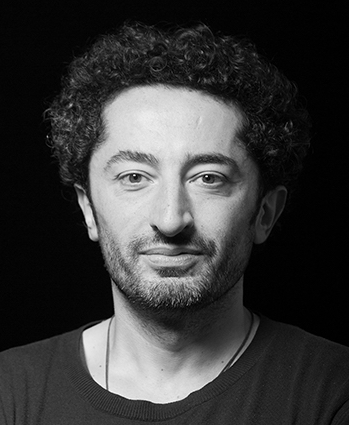}}]{Francesco Pilla}
is an Associate Professor in Smart Cities in UCD, Ireland. His work lies at the intersection between cities and technologies with the goal to build better cities through technology, innovation and citizen participation. His area of expertise is smart cities and in specific geospatial analysis and modelling of urban dynamics.
\end{IEEEbiography}

\vfill

\end{document}